\documentclass[10pt, a4paper]{article}
\usepackage{lrec2022} % this is the new LREC2022 Style
\usepackage{multibib}
\newcites{languageresource}{Language Resources}
\usepackage{graphicx}
\usepackage{tabularx}
\usepackage{soul}
\usepackage{titlesec}
\titleformat{\section}{\normalfont\large\bfseries\center}{\thesection.}{1em}{}
\titleformat{\subsection}{\normalfont\SmallTitleFont\bfseries\raggedright}{\thesubsection.}{1em}{}
\titleformat{\subsubsection}{\normalfont\normalsize\bfseries\raggedright}{\thesubsubsection.}{1em}{}
\renewcommand\thesection{\arabic{section}}
\renewcommand\thesubsection{\thesection.\arabic{subsection}}
\renewcommand\thesubsubsection{\thesubsection.\arabic{subsubsection}}
\usepackage{epstopdf}
\usepackage[utf8]{inputenc}

\usepackage{hyperref}
\hypersetup{
    colorlinks = true,
    allcolors = {blue}
}
\usepackage{xstring}

\usepackage{color}

\usepackage[acronym]{glossaries}

\makeglossaries

\newacronym{nlp}{NLP}{Natural Language Processing}
\newacronym{ner}{NER}{Named Entity Recognition}
\newacronym{sa}{SA}{Sentiment Analysis}
\newacronym{ml}{ML}{Machine Learning}
\newacronym{bow}{BoW}{bag-of-words}
\newacronym{cbow}{CBoW}{continuous Bag-of-Words}
\newacronym{sltc}{SLTC}{Swedish Language Technology Conference}
\newacronym{ann}{ANN}{artificial neural network}
\newacronym{nn}{NN}{neural network}
\newacronym{lstm}{LSTM}{Long Short Term Memory Network}
\newacronym{bilstm}{biLSTM}{bidirectional Long Short Term Memory Network}
\newacronym{sota}{SoTA}{state-of-the-art}
\newacronym{nlg}{NLG}{Natural Language Generation}
\newacronym{nlu}{NLU}{Natural Language Understanding}
\newacronym{mwe}{MWE}{Multi-Word Expression}
\newacronym{sw}{SW}{Simple Wiki}
\newacronym{mt}{MT}{Machine Translation}
\newacronym{bw}{BW}{Billion Word}
\newacronym{pie}{PIE}{Potential Idiomatic Expression}
\newacronym{iaa}{IAA}{Inter-Annotator Agreement}
\newacronym{rte}{RTE}{Recognizing Textual Entailment}
\newacronym{ir}{IR}{Information Retrieval}
\newacronym{qa}{QA}{Question Answering}
\newacronym{bnc}{BNC}{British National Corpus}
\newacronym{ukw}{UKWaC}{UK Web Pages}
\newacronym{ai}{AI}{Artificial Intelligence}
\newacronym{gdc}{GDC}{Gothenburg Dialogue Corpus}
\newacronym{dialogpt}{DialoGPT}{Dialogue Generative Pre-trained Transformer}
\newacronym{gpt}{GPT}{Generative Pre-trained Transformer}
\newacronym{multiwoz}{MultiWOZ}{Multi-Domain Wizard-of-Oz}
\newacronym{t5}{T5}{Text-to-Text Transfer Transformer}
\newacronym{bart}{BART}{Bidirectional \& Auto-Regressive Transformer}
\newacronym{xlmr}{XLM-R}{Cross-Lingual Model-RoBERTa}
\newacronym{m2m}{M2M}{Many-to-Many multilingual translation model}
\newacronym{bert}{BERT}{Bidirectional Encoder Representations from Transformers}
\newacronym{roberta}{RoBERTa}{Robustly optimized BERT pretraining Approach}
\newacronym{elmo}{ELMo}{Embeddings from Language Models}
\newacronym{pii}{PII}{personally identifiable information}
\newacronym{qg}{QG}{Question Generation}
\newacronym{tc}{TC}{Text Classification}
\newacronym{pcl}{PCL}{Patronising and Condescending Language}
\newacronym{gus}{GUS}{Genial Understander System}
\newacronym{gmb}{GMB}{Groningen Meaning Bank}
\newacronym{wsd}{WSD}{Word Sense Disambiguation}
\newacronym{ccby4}{CC-BY4}{Creative Commons Attribution 4.0}
\newacronym{ci}{CI}{confidence interval}
\newacronym{bleu}{BLEU}{bilingual evaluation understudy}
\newacronym{gdpr}{GDPR}{General Data Protection Regulation}
\newacronym{svm}{SVM}{support vector machine}
\newacronym{vs}{VS}{vector space}
\newacronym{vsm}{VSM}{vector space model}
\newacronym{nltk}{NLTK}{natural language toolkit}
\newacronym{tf-idf}{tf-idf}{term frequency-inverse document frequency}
\newacronym{pca}{PCA}{Principal Component Analysis}
\newacronym{svd}{SVD}{Singular Value Decomposition}
\newacronym{lsi}{LSI}{Latent Semantic Indexing}
\newacronym{plsi}{PLSI}{Probabilistic Latent Semantic indexing}
\newacronym{lda}{LDA}{Latent Dirichlet Allocation}
\newacronym{lm}{LM}{language model}
\newacronym{bilm}{biLM}{bidirectional language model}
\newacronym{pos}{PoS}{part of speech}
\newacronym{nnlm}{NNLM}{neural network language model}
\newacronym{bpe}{BPE}{byte-pair encoding}
\newacronym{oov}{OOV}{out-of-vocabulary}
\newacronym{imdb}{IMDB}{Internet Movie Database}
\newacronym{lr}{LR}{learning rate}
\newacronym{cus}{CUS}{Credibility unanimous score}
\newacronym{ie}{IE}{Information Extraction}
\newacronym{rl}{RL}{reinforcement learning}
\newacronym{mdl}{MDL}{minimal dependency length}
\newacronym{mlm}{MLM}{masked language model}
\newacronym{rq}{RQ}{research questions}

\title{State-of-the-art in Open-domain Conversational AI: A Survey}

\name{Tosin Adewumi*, Foteini Liwicki and Marcus Liwicki} 

\address{ML Group,   \\
         EISLAB,\\
         Luleå University of Technology, Sweden\\
         firstname.lastname@ltu.se\\}

\abstract{
We survey \acrshort{sota} open-domain conversational \acrshort{ai} models with the purpose of presenting the prevailing challenges that still exist to spur future research.
In addition, we provide statistics on the gender of conversational \acrshort{ai} in order to guide the ethics discussion surrounding the issue.
Open-domain conversational \acrshort{ai} are known to have several challenges, including bland responses and performance degradation when prompted with figurative language, among others.
First, we provide some background by discussing some topics of interest in conversational \acrshort{ai}.
We then discuss the method applied to the two investigations carried out that make up this study.
The first investigation involves a search for recent \acrshort{sota} open-domain conversational \acrshort{ai} models while the second involves the search for 100 conversational \acrshort{ai} to assess their gender.
Results of the survey show that progress has been made with recent \acrshort{sota} conversational \acrshort{ai}, but there are still persistent challenges that need to be solved, and the female gender is more common than the male for conversational \acrshort{ai}.
One main take-away is that hybrid models of conversational \acrshort{ai} offer more advantages than any single architecture.
The key contributions of this survey are 1) the identification of prevailing challenges in \acrshort{sota} open-domain conversational \acrshort{ai}, 2) the unusual discussion about open-domain conversational \acrshort{ai} for low-resource languages, and 3) the discussion about the ethics surrounding the gender of conversational \acrshort{ai}.
 \\ \newline  }

\begin{document}

\maketitleabstract

\section{Introduction}

There are different opinions as to the definition of \acrshort{ai} but according to \newcite{us2016preparing}, it is any computerised system exhibiting behaviour commonly regarded as requiring intelligence.
Conversational \acrshort{ai}, therefore, is any system with the ability to mimick human-human intelligent conversations by communicating in natural language with users \cite{jurafsky2020speech}.
Conversational \acrshort{ai}, sometimes called chatbots, may be designed for different purposes.
These purposes could be for entertainment or solving specific tasks, such as plane ticket booking (task-based).
When the purpose is to have unrestrained conversations about, possibly, many topics, then such \acrshort{ai} is called open-domain conversational \acrshort{ai}.
ELIZA, by \newcite{weizenbaum1969computer}, is the first acclaimed conversational \acrshort{ai} (or system).
Its conversations with humans demonstrated how therapeutic its responses could be.
Staff of \newcite{weizenbaum1969computer} reportedly became engrossed with the program during interactions and possibly had private conversations with it \cite{jurafsky2020speech}.

Modern \acrshort{sota} open-domain conversational \acrshort{ai} aim to achieve better performance than what was experienced with ELIZA.
There are many aspects and challenges to building such \acrshort{sota} systems.
Therefore, the primary objective of this survey is to investigate some of the recent \acrshort{sota} open-domain conversational systems and identify specific challenges that still exist that should be surmounted to achieve "human" performance in the "imitation game", as described by \newcite{turing1950computing}.
As a result of this objective, this survey will identify some of the ways of evaluating open-domain conversational \acrshort{ai}, including the use of automatic metrics and human evaluation.
This work differs from previous surveys on conversational \acrshort{ai} or related topic in that it presents discussion around the ethics of gender of conversational \acrshort{ai} with compelling statistics and discusses the uncommon topic of conversational \acrshort{ai} for low-resource languages.
Our approach surveys some of the most representative work in recent years.

The key contributions of this paper are a) the identification of existing challenges to be overcome in \acrshort{sota} open-domain conversational \acrshort{ai}, b) the uncommon discussion about open-domain conversational \acrshort{ai} for low-resource languages, and c) a compelling discussion about ethical issues surrounding the gender of conversational \acrshort{ai}.
The rest of the paper is organized as follows.
The Background Section (\ref{background}) presents brief details about some topics in conversational \acrshort{ai}; the Benefits of Conversational \acrshort{ai} Section (\ref{benefits}) highlights some of the benefits that motivate research in conversational \acrshort{ai}; the Methods Section (\ref{methods}) describes the details of the approach for the two investigations carried out in this survey; two Results of the Survey Sections (\ref{deepmodels} \& \ref{resultsethics}) then follow with details of the outcome of the methods; thereafter, the Existing Challenges Section (\ref{challenges}) shares the prevailing challenges to obtaining "human" performance; Open-domain Conversational \acrshort{ai} for Low-resource Languages Section (\ref{lowresource}) discusses this critical challenge and some of the attempts at solving it; the Related Work Section (\ref{related}) highlights previous related reviews; the Conclusion Section (\ref{conclusion}) summarizes the study after the limitations are given in the Limitation Section.

\section{Background}
\label{background}
Open-domain conversational \acrshort{ai} may be designed as a simple rule-based template system or may involve complex \acrfull{ann} architectures.
Indeed, six approaches are possible: (1) rule-based method, (2) \acrfull{rl} that uses rewards to train a policy, (3) adversarial networks that utilize a discriminator and a generator, (4) retrieval-based method that searches from a candidate pool and selects a proper candidate, (5) generation-based method that generates a response word by word based on conditioning, and (6) hybrid method \cite{jurafsky2020speech,adiwardana2020towards,chowdhary2020natural}.
Certain modern systems are still designed in the rule-based style that was used for ELIZA \cite{jurafsky2020speech}.
The \acrshort{ann} models are usually trained on large datasets to generate responses, hence, they are data-intensive.
The data-driven approach is more suitable for open-domain conversational \acrshort{ai} \cite{jurafsky2020speech}.
Such systems learn inductively from large datasets involving many turns in conversations.
A turn (or utterance) in a conversation is each single contribution from a speaker \cite{schegloff1968sequencing,jurafsky2020speech}.
The data may be from written conversations, such as the MultiWOZ \cite{eric-EtAl:2020:LREC}, transcripts of human-human spoken conversations, such as the \acrfull{gdc} \cite{allwood2003annotations}, crowdsourced conversations, such as the EmpatheticDialogues \cite{rashkin-etal-2019-towards}, and social media conversations like Familjeliv\footnote{familjeliv.se} or Reddit\footnote{reddit.com} \cite{adewumi2021sm,adewumi2022itakuroso}.
As already acknowledged that the amount of data needed for training deep \acrshort{ml} models is usually large, they are normally first pretrained on large, unstructured text or conversations before being finetuned on specific conversational data.

\subsection{Retrieval \& Generation approaches}

Two common ways that data-driven conversational \acrshort{ai} produce turns as response are \acrfull{ir} and generation \cite{jurafsky2020speech}.
In \acrshort{ir}, the system fetches information from some fitting corpus or online, given a dialogue context.
Incorporating ranking and retrieval capabilities provides additional possibilities.
If \textit{C} is the training set of conversations, given a context \textit{c}, the objective is to retrieve an appropriate turn \textit{r} as the response.
Similarity is used as the scoring metric and the highest scoring turn in \textit{C} gets selected from a potential set.
This can be achieved with different \acrshort{ir} methods
%including the classic tf-idf for \textit{C} and \textit{c},
and choosing the response with the highest cosine similarity with \textit{c} \cite{jurafsky2020speech}.
This is given in Equation~\ref{eqir}.
%A neural \acrshort{ir} method is another approach one could use.
In an encoder-encoder architecture, for example, one could train the first encoder to encode the query while the second encoder encodes the candidate response and the score is the dot product between the two vectors from both encoders.
In the generation method, a language model or an encoder-decoder is used for response generation, given a dialogue context.
As shown in Equation~\ref{eqgen}, each token of the response ($r_t$) of the encoder-decoder model is generated by conditioning on the encoding of the query ($q$) and all the previous responses ($r_{t-1}...r_1$), where $w$ is a word in the vocabulary $V$.
Given the benefit of these two methods, it may be easy to see the advantage of using the hybrid of the two for conversational \acrshort{ai}.

\begin{equation}
\label{eqir}
    response (\textit{c,C}) = \arg\max_{r \epsilon C } \frac{c.r}{|c||r|}
\end{equation}

\begin{equation}
\label{eqgen}
    r_t = \arg\max_{w \epsilon V } P(w|q,r_{t-1}...r_1)
\end{equation}

\subsection{Evaluation}

Although there are a number of metrics for \acrshort{nlp} systems \cite{aggarwal2012survey,gehrmann-etal-2021-gem,reiter201020} different metrics may be suitable for different systems, depending on the characteristics of the system.
For example, the goals of task-based systems are different from those of open-domain conversational systems, so they may not use the same evaluation metrics.
Human evaluation is the \textit{gold standard} in the evaluation of open-domain conversational \acrshort{ai}, though it is subjective \cite{zhang2020dialogpt}.
It is both time-intensive and laborious.
As a result of this, automatic metrics serve as proxies for estimating performance though they may not correlate very well with human evaluation \cite{gehrmann-etal-2021-gem,gangal-etal-2021-improving,jhamtani-etal-2021-investigating}.
For example, \acrshort{ir} systems may use F1, precision, and recall \cite{aggarwal2012survey}.
Furthermore, metrics used in \acrshort{nlg} tasks, like \acrfull{mt}, such as the \acrshort{bleu} or ROUGE, are sometimes used to evaluate conversational systems \cite{zhang2020dialogpt} but they are also discouraged because they do not correlate well with human judgment \cite{liu2016not,jurafsky2020speech}.
They do not take syntactic or lexical variation into consideration \cite{reiter201020}.
Dependency-based evaluation metrics, however, allow for such variation in evaluation.
Perplexity is commonly used for evaluation and has been shown to correlate with a human evaluation metric called Sensibleness and Specificity Average  (SSA) \cite{adiwardana2020towards}.
It measures how well a model predicts the data of the test set, thereby estimating how accurately it expects the words that will be said next \cite{adiwardana2020towards}.
It corresponds to the effective size of the vocabulary \cite{aggarwal2012survey} and smaller values show that a model fits the data better.
Very low perplexity, however, has been shown to suggest such text may have low diversity and unnecessary repetition \cite{holtzman2019curious}.

Two methods for human evaluation of open-domain conversational \acrshort{ai} are observer and participant evaluation \cite{jurafsky2020speech}.
Observer evaluation involves reading and scoring a transcript of human-chatbot conversation while participant evaluation interacts directly with the conversational \acrshort{ai} \cite{jurafsky2020speech}.
In the process, the system may be evaluated for different qualities, such as humanness (or human-likeness), fluency, making sense, engagingness, interestingness, avoiding repetition, and more.
The Likert scale is usually provided for grading these various qualities.
The others are comparison of diversity and how fitting responses are to the given contexts.
Human evaluation is usually modeled to resemble the Turing test (or the imitation game).

\subsection{The Turing test}
Modern human evaluation is generally designed like the Turing test. 
The Turing test is the indistinguishability test.
This is when a human is not able to distinguish if the responses are from another human or a machine in what is called the imitation game \cite{turing1950computing}.
The proposed imitation game, by \newcite{turing1950computing}, involves a man, a woman, and an interrogator of either sex, who is in a separate room from the man and the woman.
The goal of the interrogator is to determine who is the woman and who is the man, and he does this by directing questions to the man and the woman, which are answered in some written format.
The man tries to trick the interrogator into believing he's a woman while the woman tries to convince the interrogator she's a woman.
When a machine replaces the man, the aim is to find out if the interrogator will decide wrongly as often as when it was played with a man \cite{turing1950computing}.
The formulation of the imitation game, by \newcite{turing1950computing}, does not precisely match modern versions of the test \cite{saygin2002pragmatics}.

%The Turing test has different versions \citep{traiger2003making}.
%Indeed, at some point in the same paper by \cite{turing1950computing}, after replacing the man with a machine, the woman is also replaced by a man.

An early version of the test was applied to PARRY, a chatbot designed by \newcite{colby1972turing} to imitate aggressive emotions.
Most psychiatrists couldn't distinguish between transcripts of real paranoids and PARRY \cite{colby1971artificial,jurafsky2020speech}.
%However, this is disputed by some, since ELIZA was able to fool many of its users as well \citep{mauldin1994chatterbots,jurafsky2020speech}.
This example of PARRY may be viewed as an edge case, given that the comparison was not made with rational human beings but paranoids \cite{mauldin1994chatterbots}.
A limited version of the test was introduced in 1991, alongside its unrestricted version, in what is called the Loebner Prize competition \cite{mauldin1994chatterbots}.
Every year, since then, prizes have been awarded to conversational \acrshort{ai} that pass the restricted version in the competitions \cite{bradevsko2012survey}.
This competition has its share of criticisms, including the view that it is rewarding tricks instead of furthering the course of \acrshort{ai} \cite{shieber1994lessons,mauldin1994chatterbots}.
As a result of this, \newcite{shieber1994lessons} recommended an alternative approach, whereby the competition will involve a different award methodology that is based on a different set of assessment and done on an occasional basis.

\subsection{Characteristics of Human Conversations}

Humans converse using speech and other gestures that may include facial expressions, usually called body language, thereby making human conversations complex \cite{jurafsky2020speech}.
Similar gestures may be employed when writing conversations.
Such gestures may be clarification questions or the mimicking of sound (\textit{onomatopoeia}).
In human conversations, one speaker may have the conversational initiative, i.e., the speaker directs the conversation.
This is typical in an interview where the interviewer asking the questions directs the conversation.
It is the style for \acrfull{qa} conversational \acrshort{ai}.
In typical human-human conversations, the initiative shifts to and from different speakers.
This kind of mixed (or rotating) initiative is harder to achieve in conversational systems \cite{jurafsky2020speech}.
Besides conversation initiative, below are additional characteristics of human conversations, according to \newcite{sacks1978simplest}.

\begin{itemize}
    \item Usually, one speaker talks at a time.
    \item The turn order varies.
    \item The turn size varies.
    \item The length of a conversation is not known in advance.
    \item The number of speakers/parties may vary.
    \item Techniques for allocating turns may be used.
\end{itemize}

\subsection{Ethics}
Ethical issues are important in open-domain conversational \acrshort{ai}.
And the perspective of deontological ethics views objectivity as being equally important \cite{adewumi2019conversational,javed2021understanding,white2009immanuel}.
Deontological ethics is a philosophy that emphasizes duty or responsibility over the outcome achieved in decision-making \cite{alexander2007deontological,paquette2015ends}.
Responsible research in conversational \acrshort{ai} requires compliance to ethical guidelines or regulations, such as the \acrfull{gdpr}, which is a regulation protecting persons with regards to their personal data \cite{voigt2017eu}.
Some of the ethical issues that are of concern in conversational \acrshort{ai} are privacy, due to \acrfull{pii}, toxic/hateful messages as a result of the training data and unwanted bias (racial, gender, or other forms) \cite{jurafsky2020speech,zhang2020dialogpt}.

Some systems have been known to demean or abuse their users.
It is also well known that machine learning systems reflect the biases and toxic content of the data they are trained on \cite{neff2016automation,jurafsky2020speech}.
Privacy is another crucial ethical issue.
Data containing \acrshort{pii} may fall into the wrong hands and cause security threat to those concerned.
It is important to have systems designed such that they are robust to such unsafe or harmful attacks.
Attempts are being made with debiasing techniques to address some of these challenges \cite{dinan-etal-2020-queens}.
Privacy concerns are also being addressed through anonymisation techniques \cite{henderson2018ethical,jurafsky2020speech}.
Balancing the features of chatbots with ethical considerations can be a delicate and challenging work.
For example, there is contention in some quarters whether using female voices in some technologies/devices is appropriate.
Then again, one may wonder if there is anything harmful about that.
This is because it seems to be widely accepted that the proportion of chatbots designed as “female” is larger than the those designed as “male”.
In a survey of 1,375 chatbots, from automatically crawling chatbots.org, \newcite{maedche2020gender} found that most were female.

\section{Benefits of Conversational \acrshort{ai}}
\label{benefits}

The apparent benefits inherent in open-domain conversational \acrshort{ai} has spurred research in the field since the early days of ELIZA.
These benefits have led to investments in conversational \acrshort{ai} by many organizations, including Apple \cite{jurafsky2020speech}.
Some of the benefits include:
\begin{itemize}
    \item Provision of therapeutic company, as was experienced with ELIZA.
    \item The provision of human psychiatric/psychological treatment on the basis of favorable behavior determined from experiments which are designed to modify input-output behaviour in models.
    This may be designed like PARRY \cite{colby1971artificial}.
    \item Provide support for users with disabilities, such as blindness \cite{reiter201020}.
    \item A channel for providing domain/world knowledge \cite{reiter201020}.
    \item The provision of educational content or information in a concise fashion \cite{kerry2008conversational}.
    \item Automated machine-machine generation of quality data for low-resource languages \cite{adewumi2022itakuroso}.
\end{itemize}

\section{Methods}
\label{methods}
\begin{figure}[!h]
\begin{center}
\includegraphics[width=0.5\textwidth]{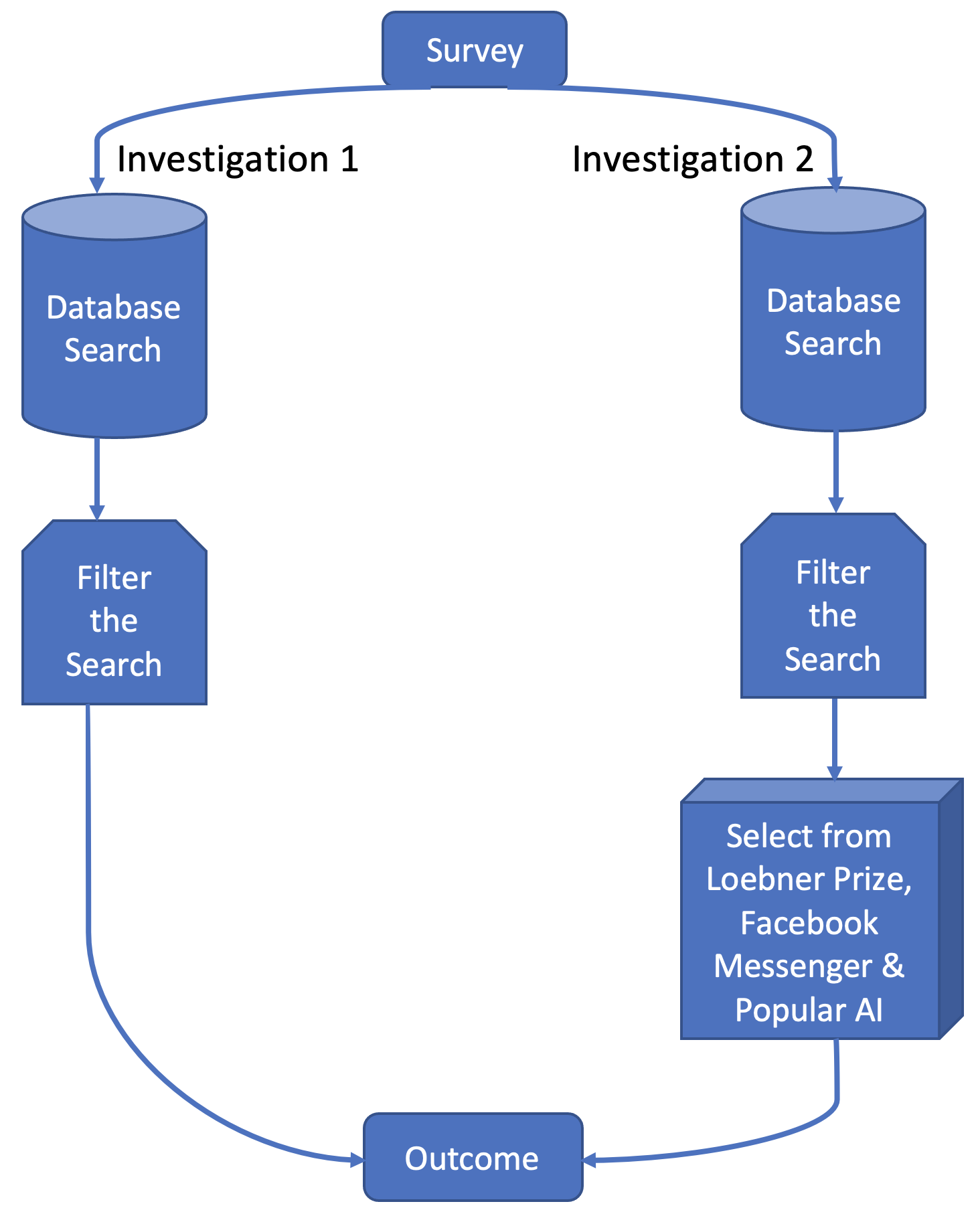}
\caption{Method for both investigations in this study}
\label{fig1}
\end{center}
\end{figure}

We conduct two different investigations to make up this survey.
Figure~\ref{fig1} depicts the methods for both investigations.
The first addresses text-based, open-domain conversational \acrshort{ai} in terms of architectures while the second addresses the ethical issues about the gender of such systems.
The first involves online search on Google Scholar and regular Google Search, using the term "\acrlong{sota} open-domain dialogue systems".
This returned 5,130 and 34,100,000 items in the results for Google Scholar and Google Search, respectively. 
We then sieve through the list of scientific papers (within the first ten pages because of time-constraint) to identify those that report \acrshort{sota} results in the last five years (2017-2022) in order to give more attention to them.
It is important to note that some Google Scholar results point to other databases, like ScienceDirect, arXiv, and IEEEXplore.
The reason for also searching on regular Google Search is because it provides results that are not necessarily based on peer-reviewed publications but may be helpful in leading to peer-reviewed publications that may not have been immediately obvious on Scholar.
We did not discriminate the papers based on the field of publication, as we are interested in as many \acrshort{sota} open-domain conversational systems as possible, within the specified period.
A second stage involves classifying, specifically, the \acrshort{sota} open-domain conversational \acrshort{ai} from the papers, based on their architecture.
We also consider models that are pretrained on large text and may be adapted for conversational systems, such as the \acrfull{t5} \cite{JMLR:v21:20-074}, and autoregressive models because they easily follow the \acrshort{nlg} framework.
We do not consider models for which we did not find their scientific papers.
%conversational \acrshort{ai} are discussed in Section \ref{deepmodels}.

The second investigation, which addresses the ethical issues surrounding the gender of conversational \acrshort{ai}, involves the survey of 100 chatbots.
It is based on binary gender: male and female.
The initial step was to search using the term "gender chatbot" on Google Scholar and note all chatbots identified in the scientific papers in the first ten pages of the results.
Then, using the same term, the Scopus database was queried and it returned 20 links.
The two sites resulted in 120 links, from which 59 conversational systems were identified.
Since Facebook Messenger is linked to the largest social media platform, we chose this to provide another 20 chatbots.
They are based on information provided by two websites on some of the best chatbots on the platform\footnote{enterprisebotmanager.com/chatbot-examples \\ growthrocks.com/blog/7-messenger-chatbots}.
The sites were identified on Google by using the search term "Facebook Messenger best chatbots".
They were selected based on the first to appear in the list.
To make up part of the 100 conversational \acrshort{ai}, 13 chatbots, which have won the Loebner prize in the past 20 years, are included in this survey.
Finally, 8 popular conversational \acrshort{ai}, which are also commercial, are included.
These are Microsoft’s XiaoIce and Cortana, Amazon's Alexa, Apple's Siri, Google Assistant, Watson Assistant, Ella, and Ethan by Accenture.

\section{Results of Survey: Models}
\label{deepmodels}

Review of the different scientific papers from the earlier method show that recent \acrshort{sota} open-domain conversational \acrshort{ai} models fall into one of the latter three  approaches mentioned in Section \ref{background}: (a) retrieval-based, (b) generation-based, and (c) hybrid approaches.
The models are BlenderBot 1 \& 2, Meena, DLGNet, \acrfull{dialogpt}, \acrfull{gpt}-3 and 2 (RetGen), and \acrfull{t5}.

\subsection{BlenderBot 1 \& 2}

Some of the ingredients for the success of BlenderBot, as identified by \newcite{roller2020recipes}, are empathy and personality, consistent persona, displaying knowledge, and engagingness.
Three different parameter models are built for the variants: 90M, 2.7B, and 9.4B.
The variants, which are all based on the Transformer, involve the latter three approaches: retrieval, generative, and a retrieve-and-refine combination of the earlier two.
The generative architecture is a seq2seq model and uses Byte-Level \acrshort{bpe} for tokenization.
Human evaluation of multi-turn conversations, using ACUTE-Eval method, shows that its best model outperforms the previous \acrshort{sota} on engagingness and humanness by using the Blended Skill Talk (BST) dataset \cite{smith-etal-2020-put}.
They observed that models may give different results when different decoding algorithms are used, though the models may report the same perplexity in automatic metric.
The more recent version of the set of models learns to generate an online search query from the internet based on the context and conditions on the results to generate a response, thereby employing the latest relevant information \cite{komeili2021internet,xu2021beyond}.

The seq2seq (or encoder-decoder) is an important standard architecture for BlenderBot and other conversational \acrshort{ai} \cite{xu2021beyond}.
The Transformer, by \cite{vaswani2017attention}, is often used as the underlying architecture for it, though the \acrfull{lstm}, by \newcite{hochreiter1997long}, may also be used.
Generally, the encoder-decoder conditions on the encoding of the prompts and responses up to the last time-step for it to generate the next token as response \cite{adiwardana2020towards,jurafsky2020speech}.
The sequence of tokens is run through the encoder stack's embedding layer, which then compresses it in the dense feature layer into fixed-length feature vector.
A sequence of tokens is then produced by the decoder after they are passed from the encoder layer.
The Softmax function is then used to normalized this, such that the token with the highest probability is the output.
%Attention \citep{bahdanau2015neural} may be introduced to the model.
%The attention mechanism focuses on desired parts of a sequence regardless of where they may appear in the input and ignores other parts or assigns less weighted average to them \citep{JMLR:v21:20-074}.

\subsection{Meena}
Meena is presented by \newcite{adiwardana2020towards}.
It is a multi-turn open-domain conversational \acrshort{ai} seq2seq model that was trained end-to-end \cite{bahdanau2015neural}.
The underlying architecture of this seq2seq model is the Evolved Transformer (ET).
It has 2.6B parameters and includes 1 ET encoder stack and 13 ET decoder stacks.
Manual coordinate-descent search was used to determine the hyperparameters of the best Meena model.
The data it was trained on is a filtered public domain corpus of social media conversations containing 40B tokens.
Perplexity was used to automatically evaluate the model.
It was also evaluated in multi-turn conversations using the human evaluation metric: Sensibleness and Specificity Average (SSA).
This combines two essential aspects of a human-like chatbot: being specific and making sense.

\subsection{DLGNet}
DLGNet is presented by \newcite{olabiyi2019multiturn}.
Its architecture is similar to \acrshort{gpt}-2, being an autoregressive model.
It is a multi-turn dialogue response generator that was evaluated, using the automatic metrics \acrshort{bleu}, ROUGE, and distinct n-gram, on the Movie Triples and closed-domain Ubuntu Dialogue datasets.
It uses multiple layers of self-attention to map input sequences to output sequences.
This it does by shifting the input sequence token one position to the right so that the model uses the previously generated token as additional input for the next token generation.
Given a context, it models the joint distribution of the context and response, instead of modeling the conditional distribution.
Two sizes were trained: a 117M-parameter model and the 345M-parameter model.
The 117M-parameter model has 12 attention layers while the 345M-parameter model has 24 attention layers.
The good performance of the model is attributed to the long-range transformer architecture, the injection of random informative paddings, and the use of  \acrshort{bpe}, which provided 100\% coverage for Unicode texts and prevented the \acrshort{oov} problem.

\subsection{\acrshort{dialogpt} 1 \& 2 (RetGen)}
\acrshort{dialogpt} was trained on Reddit conversations of 147M exchanges \cite{zhang2020dialogpt}.
It is an autoregressive \acrshort{lm} based on \acrshort{gpt}-2.
Its second version (RetGen) is a hybrid retrieval-augmented/grounded version.
In single-turn conversations, it achieved \acrshort{sota} in human evaluation and performance that is close to human in open-domain dialogues, besides achieving \acrshort{sota} in automatic evaluation.
The large model has 762M parameters with 36 Transformer layers; the medium model has 345M parameters with 24 layers; the small model has 117M parameters with 12 layers.
A multiturn conversation session is framed as a long text in the model and the generation as language modeling.
%It employs what is called maximum mutual information (MMI) scoring to address the problem of bland response.
%This technique uses a pretrained backward model to the source sentences from the responses.
The model is easily adaptable to new dialogue datasets with few samples.
The RetGen version of the model jointly trains a grounded generator and document retriever \cite{zhang2021joint}.

\subsection{\acrshort{gpt}-3 \& \acrshort{gpt}-2}
\acrshort{gpt}-3 is introduced by \newcite{brown2020language}, being the largest size out of the eight models they created.
It is a 175B-parameter autoregressive model that shares many of the qualities of the \acrshort{gpt}-2 \cite{radford2019language}.
These include modified initialization, reversible tokenization, and pre-normalization.
However, it uses alternating dense and locally banded sparse attention.
Both \acrshort{gpt}-3 and \acrshort{gpt}-2 are trained on the CommonCrawl dataset, though different versions of it.
\acrshort{gpt}-3 achieves strong performance on many \acrshort{nlp} datasets, including open-domain \acrshort{qa}.
In addition, zero-shot perplexity, for automatic metric, was calculated on the Penn Tree Bank (PTB) dataset.
Few-shot inference results reveal that it achieves strong performance on many tasks.
Zero-shot transfer is based on providing text description of the task to be done during evaluation.
It is different from one-shot or few-shot transfer, which are based on conditioning on 1 or k number of examples for the model in the form of context and completion.
No weights are updated in any of the three cases at inference time and there's a major reduction of task-specific data that may be needed.

\subsection{\acrshort{t5}}
\acrshort{t5} was introduced by \newcite{JMLR:v21:20-074}.
It is an encoder-decoder Transformer architecture and has a multilingual version, m\acrshort{t5} \cite{xue-etal-2021-mt5}.
It is trained on Colossal Clean Crawled Corpus (C4) and achieved \acrshort{sota} on the SQuAD \acrshort{qa} dataset, where it generates the answer token by token.
A simplified version of layer normalization is used such that no additive bias is used, in contrast to the original Transformer.
The self-attention of the decoder is a form of causal or autoregressive self-attention.
All the tasks considered for the model are cast into a unified text-to-text format, in terms of input and output.
This approach, despite its benefits, is known to suffer from prediction issues \cite{adewumi2022ml_ltu,sabry2022hat5}.
Maximum likelihood is the training objective for all the tasks and a task-prefix is specified in the input before the model is fed, in order to identify the task at hand. The base version of the model has about 220M parameters.

\section{Results \& Discussion of Survey: Ethics of Gender}
\label{resultsethics}

\begin{figure*}[!h]
\begin{center}
\includegraphics[width=1\textwidth]{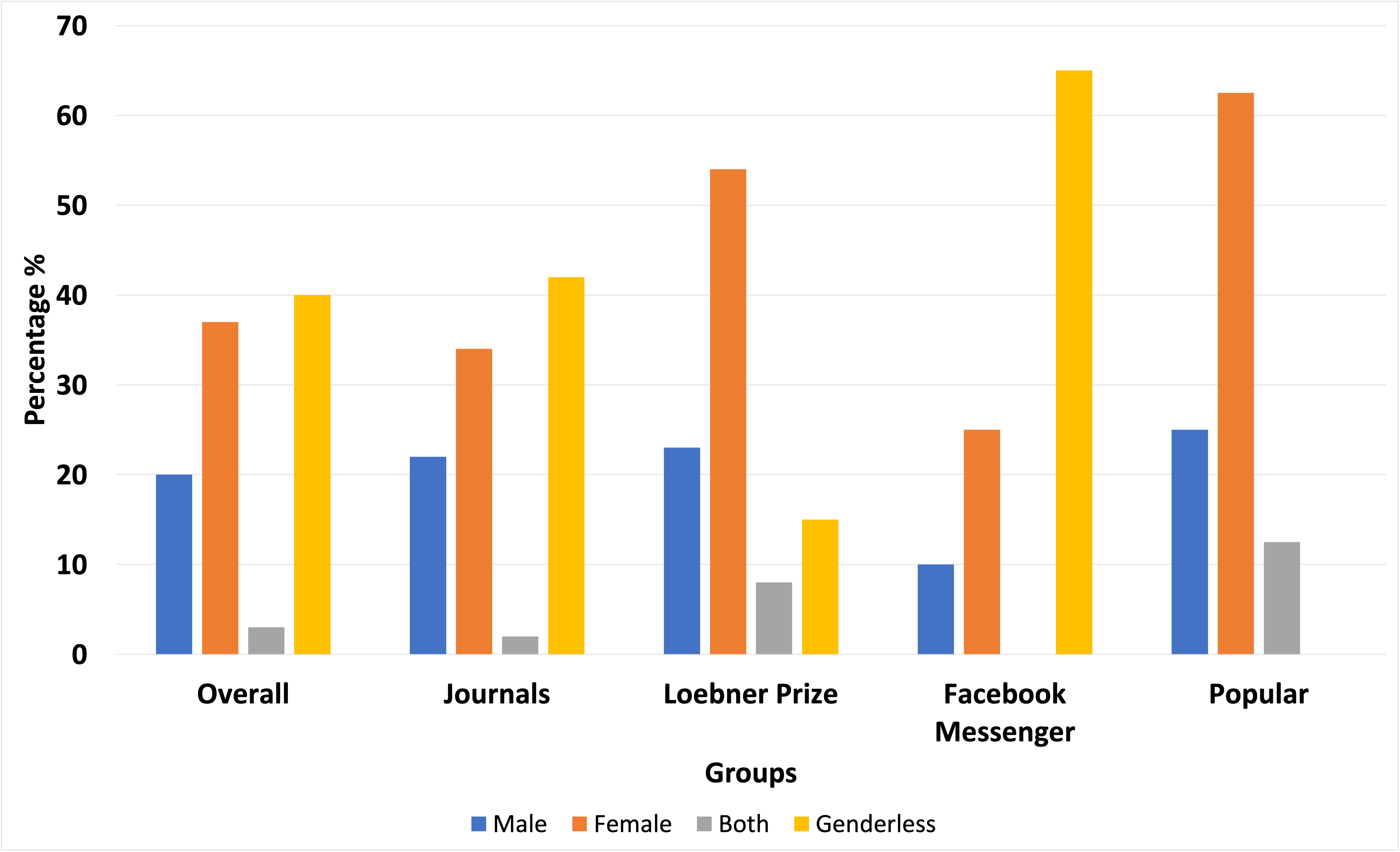}
\caption{Method for both investigations in this study}
\label{fig2}
\end{center}
\end{figure*}

Following the procedure mentioned in Section \ref{methods}, each conversational \acrshort{ai}'s gender is determined by the designation given by the developer or cues such as avatar, voice or name, for cases where the developer did not identify the gender.
These cues are based on general perception or stereotypes.
We consider a conversational \acrshort{ai} genderless if it is specifically stated by the reference or developer or nothing is mentioned about it and there are no cues to suggest gender.
%\cite{maedche2020gender} uses similar cues in their research.
Overall, in the investigation of the 100 conversational \acrshort{ai}, 37 (or 37\%) are female, 20 are male, 40 are genderless, and 3 have both gender options.
Figure \ref{fig2} shows a bar graph with details of the results.
Breaking down the data into 4 groups: journal-based, Loebner-winners, Facebook Messenger-based, and popular/commercial chatbots, we observe that female conversational \acrshort{ai} always outnumber male conversational \acrshort{ai}.
The genderless category does not follow such a consistent trend in the groups.
Out of the 59 chatbots mentioned in journal articles, 34\% are female, 22\% are male, 42\% are genderless, and 2\% have both gender options.
54\% are female among the 13 chatbots in the Loebner-winners, 23\% are male, 15\% are genderless, and 8\% have both options. Of the 20 chatbots from Facebook Messenger, 25\% are female, 10\% are male, 65\% are genderless, and 0 offer both genders.
Lastly, out of the 8 popular/commercial conversational \acrshort{ai}, 62.5\% are female, 25\% are male, 0 is genderless, and 12.5\% have both options.

\subsection{Discussion}

The results agree with the popular assessment that female conversational \acrshort{ai} are more predominant than the male ones.
We do not know of the gender of the producers of these 100 conversational \acrshort{ai} but it may be a safe assumption that most are male.
This assessment has faced criticism from some interest groups, evidenced in a recent report by \newcite{west2019d} that the fact that most conversational \acrshort{ai} are female makes them the face of glitches resulting from the limitations of \acrshort{ai} systems.
Despite the criticism, there's the opinion that this phenomenon can be viewed from a vantage position for women.
For example, they may be viewed as the acceptable face, persona or voice, as the case may be, of the planet.
A comparison was made by \newcite{silvervarg2012effect} of a visually androgynous agent with both male and female agents and it was found that it suffered verbal abuse less than the female agent but more than the male agent.
Does this suggest developers do away with female conversational \acrshort{ai} altogether to protect the female gender or what is needed is a change in the attitude of users?
Especially since previous research has shown that stereotypical agents, with regards to task, are often preferred by users \cite{forlizzi2007interface}.
Some researchers have argued that conversational \acrshort{ai} having human-like characteristics, including gender, builds trust for users \cite{louwerse2005social,muir1987trust,nass2005wired}.
Furthermore, \newcite{lee2019caring} observed that conversational \acrshort{ai} that consider gender of users, among other cues, are potentially helpful for self-compassion of users.
Noteworthy that there are those who consider the ungendered, robotic voice of \acrshort{ai} uncomfortable and eerie and will, therefore, prefer a specific gender.

\section{Existing Challenges of Open-domain Conversational \acrshort{ai}}
\label{challenges}

This survey has examined several \acrshort{sota} open-domain conversational \acrshort{ai} models.
Despite their noticeable successes and the general progress, challenges still remain.
The challenges contribute to the non-human-like utterances the conversational \acrshort{ai} tend to have.
These challenges also provide motivation for active research in \acrshort{nlp}.
For example, the basic seq2seq architecture is known for repetitive and dull responses \cite{chowdhary2020natural}.
One way of augmenting the architecture for refined responses is the use of \acrshort{ir} techniques, like concatenation of retrieved sentences from Wikipedia to the conversation context \cite{jurafsky2020speech}.
Other shortcomings may be handled by switching the objective function to a mutual information objective or introducing the beam search decoding algorithm in order to achieve relatively more diverse responses \cite{chowdhary2020natural}.
Besides, \acrshort{gpt}-3 is observed to lose coherence over really long passages, gives contradictory utterances, and its size is so large that it's difficult to deploy.
Collectively, some of the existing challenges are highlighted below.
It is hoped that identifying these challenges will spur further research in these areas.

\begin{enumerate}
    \item Poor coherence in sequence of text or across multiple turns of generated conversation \cite{jurafsky2020speech,welleck2019neural}.
    \item Lack of utterance diversity \cite{holtzman2019curious}.
    \item Bland repetitive utterances \cite{holtzman2019curious,zhang2020dialogpt}.
    \item Lack of empathetic responses from conversational systems \cite{rashkin-etal-2019-towards}.
    \item Lack of memory to personalise user experiences.
    \item Style inconsistency or lack of persona \cite{adiwardana2020towards,zhang2020dialogpt}.
    \item Multiple initiative coordination \cite{jurafsky2020speech}.
    \item Poor inference and implicature during conversation.
    \item Lack of world-knowledge.
    \item Poor adaptation or responses to idioms or figurative language \cite{jhamtani-etal-2021-investigating}
    \item Hallucination of facts when generating responses \cite{marcus2018deep}.
    \item Obsolete facts, which are frozen in the models' weights during at training .
    \item Training requires a large amount of data \cite{marcus2018deep}.
    \item Lack of common-sense reasoning \cite{marcus2018deep}.
    \item Large models use so many parameters that make them complex and may impede transparency \cite{marcus2018deep}.
    \item Lack of training data for low-resource languages \cite{adewumi2022itakuroso,adewumi2020challenge}

\end{enumerate}

\section{Open-domain Conversational \acrshort{ai} for Low-resource Languages}
\label{lowresource}
The last challenge mentioned in the earlier section is a prevailing issue for many languages around the world.
Low-resource languages are natural languages with little or no digital data or resources \cite{nekoto-etal-2020-participatory,adewumi2022itakuroso}.
This challenge has meant that so many languages are unrepresented in many deep \acrshort{ml} models, as they usually require a lot of data for pretraining.
Even 
Noteworthy, though, that multilingual versions of some of the models are being made with very limited data of the low-resource languages.
They are, however, known to have relatively poor performance compared to models trained completely on the target language data \cite{pfeiffer2020AdapterHub,virtanen2019multilingual,ronnqvist2019multilingual} and only few languages are covered \cite{adewumi2022itakuroso}.
Approaches to mitigating this particular challenge involve human and automatic \acrshort{mt} attempts  \cite{nekoto-etal-2020-participatory} and efforts at exploiting cross-lingual transfer to build conversational \acrshort{ai} capable of machine-machine conversations for automated data generation \cite{adewumi2022itakuroso}.

\section{Related Work}
\label{related}
In a recent survey, \newcite{caldarini2022literature} reviewed advances in chatbots by using the common approach of acquiring scientific papers from search databases, based on certain search terms, and selecting a small subset from the lot for analysis, based on publications between 2007 and 2021.
The databases they used are IEEE, ScienceDirect, Springer, Google Scholar, JSTOR, and arXiv.
They analyzed rule-based and data-driven chatbots from the filtered collection of papers.
Their distinction of rule-based chatbots as being different from \acrshort{ai} chatbots may be disagreed with, especially when a more general definition of \acrshort{ai} is given and since  modern systems like Alexa have rule-based components \cite{jurafsky2020speech}.
Meanwhile, \newcite{fu2022learning} reviewed learning towards conversational \acrshort{ai} and in their survey classified conversational \acrshort{ai} into three frameworks.
They posit that a human-like conversation system should be both (1) informative and (2) controllable.

A systematic survey of recent advances in deep learning-based dialogue systems was conducted by \newcite{ni2021recent}, where the authors recognise that dialogue modelling is a complicated task because it involves many related \acrshort{nlp} tasks, which are also required to be solved.
They categorised dailogue systems by analysing them from two angles: model type and system type (including task-oriented and open-domain conversational systems).
\newcite{khatri2018advancing} also recognised that building open-domain conversational \acrshort{ai} is a challenging task.
They describe how, through the Alexa Prize, teams advanced the \acrshort{sota} through context in dialog models, using knowledge graphs for language understanding, and building statistical
and hierarchical dialog managers, among other things.

\section{Limitation}
Although this work has presented recent \acrshort{sota} open-domain conversational \acrshort{ai} within the first ten pages of the search databases (Google Scholar \& Google Search) that were used, we recognise that the time-constraint and restricted number of pages of results means there may have been some that were missed.
This goes also for the second investigation on the gender of conversational \acrshort{ai}.
Furthermore, our approach did not survey all possible methods for conversational \acrshort{ai}, though it identified all the major methods available.

\section{Conclusion}
\label{conclusion}
In this survey of the \acrshort{sota} open-domain conversational \acrshort{ai}, we identified models that have pushed the envelope in recent times.
It appears that hybrid models of conversational \acrshort{ai} offer more advantages than any single architecture, based on the benefit of up-to-date responses and world knowledge.
Besides discussing some of their successes or strengths, we focused on prevailing challenges that still exist and which need to be surmounted to achieve the type of desirable performance, typical of human-human conversations.
The important challenge with conversational \acrshort{ai} for low-resource languages is highlighted and the ongoing attempts at tackling it.
The presentation of the discussion on the ethics of the gender of conversational \acrshort{ai} gives a balanced perpective to the debate.
We believe this survey will spur focused research in addressing some of the challenges identified, thereby enhancing the \acrshort{sota} in open-domain conversational \acrshort{ai}.

% \nocite{*}
\section{Bibliographical References}\label{reference}
%\label{main:ref}

\bibliographystyle{lrec2022-bib}
\bibliography{lrec2022-example}

%\section{Language Resource References}
%\label{lr:ref}
%\bibliographystylelanguageresource{lrec2022-bib}
%\bibliographylanguageresource{languageresource}

\printglossary[type=\acronymtype]

\end{document}